**Generative AI Expands the Intellectual Reach of Course Based Undergraduate Research Experiences (CUREs)**

Aditi Babar[1], Kristin J. Davin[2] , Alex Dornburg[1*]

[1]*Department of Bioinformatics and Genomics, University of North Carolina at Charlotte; Charlotte, NC, USA*

[2] *Department of Secondary, Middle, and K-12 Education, University of North Carolina at Charlotte, NC, USA*

**ORCID IDs**:

K. Davin: 0000-0003-3590-7086
A. Dornburg: 0000-0003-0863-2283

*** Corresponding author at:**
Alex Dornburg, Department of Bioinformatics and Genomics, University of North Carolina at Charlotte, Charlotte, North Carolina, USA. E-mail address: adornbur@charlotte.edu. Phone: +1 704-687-8437

**Acknowledgements.** We thank members of the Dornburg laboratory group for providing feedback during the early stages of this work.

**Funding.** This work was supported by the National Science Foundation (NSF, United States) under award IOS2419128 to AD and KD and institutional support from the University of North Carolina at Charlotte. The content is solely the responsibility of the authors and does not necessarily represent the official views of the NSF or UNC Charlotte.

**Abstract**

Course-based undergraduate research experiences (CUREs) broaden access to authentic scientific inquiry through responsive instructor support as research problems become increasingly complex. Generative artificial intelligence (GenAI) may extend this support by providing individualized assistance that can adapt as student needs change. However, how embedding GenAI within a CURE to provide support across the research process impacts student inquiry, collaboration, and scientific reasoning remains unresolved. Here we use longitudinal qualitative data collected across three semesters of a bioinformatics and genomics CURE to show that GenAI expanded the intellectual reach of the research experience in three distinct ways. First, personalized, on-demand scaffolding allowed students to move beyond the boundaries of instructor expertise and transform their own interests into researchable inquiry, with all teams developing distinct self-directed projects rather than selecting instructor-provided topics. Second, GenAI became part of the distributed cognitive system of research teams, helping novice researchers communicate and coordinate across differentiated expertise without eliminating specialization. Third, expanded capability did not replace the need for disciplinary judgment. Students increasingly validated, revised, or rejected AI-generated contributions, such that research independence emerged through retained intellectual responsibility. Together, these findings suggest that GenAI can extend the reach of CUREs by expanding what novice researchers can investigate, how they can collaborate, and the level of responsibility they can assume while preserving human judgment central to authentic scientific inquiry.

## 1. Introduction

Course-based undergraduate research experiences (CUREs) have emerged as an important strategy for expanding participation in authentic scientific research within the undergraduate curriculum (Auchincloss et al. 2014; Bangera and Brownell 2014). By relocating authentic research from predominantly apprenticeship-based settings into the classroom, CUREs support the early development of research skills, scientific self-efficacy, and scientific identity while promoting persistence in STEM fields (Shaffer et al. 2010; Rodenbusch et al. 2016; Cooper et al. 2019). However, implementing CUREs requires sustained instructional and institutional support. Authentic inquiry depends on responsive feedback and instructors capable of recognizing and addressing the evolving needs of students in research settings (Goodwin et al. 2022; Cooper et al. 2026), consequently placing substantial demands on instructor preparation and mentoring capacity (Spell et al. 2014; Craig 2017; DeChenne-Peters and Scheuermann 2022). As more institutions look to develop CUREs, these needs also place new training demands on other instructional personnel such as teaching assistants (Heim and Holt 2019; Kern and Olimpo 2023; Santillan et al. 2024) and often require institutional and fiscal support (Lopatto et al. 2014; Bell et al. 2017). These challenges become increasingly consequential as CUREs expand across larger student populations, necessitating the evaluation of complementary strategies that support authentic inquiry for learners.

The rise of generative artificial intelligence (GenAI) tools offers a novel means of extending support to students engaged in complex, open-ended learning. Unlike static instructional resources, large language models can respond dynamically to individual questions, provide immediate and iterative feedback, generate alternative explanations, and help students refine ideas or approaches as their understanding develops (Johnson et al. 2024; Abdallah et al. 2025; Qian 2025). Such capabilities may be particularly valuable in CURE environments, where developing research competency requires students to formulate questions, interpret evidence, and progressively assume greater responsibility for scientific decision-making. Moreover, GenAI can create on-demand opportunities for instruction tailored to the level of the learner (Huang et al. 2023; Kestin et al. 2025), while also reshaping how students generate ideas (Habib et al. 2024) and engage with peers during collaborative learning (Daniel et al. 2025; Wei and Perkins 2026). However, GenAI can also undermine learning when students rely on these tools to perform the cognitive work that instructional activities are intended to develop. Unstructured use of GenAI can encourage cognitive offloading and overreliance, potentially diminishing information retention and critical engagement (Zhai et al. 2024; Poulidis et al. 2025; Qian 2025). In the worst-case scenario, unrestricted use of GenAI for outsourcing learning tasks has been shown to impede subsequent learning when access to these tools is restricted (Bastani et al. 2025). Determining whether GenAI strengthens or weakens CUREs therefore requires examining how students engage with these tools as sources of personalized support, collaborative scaffolding, and guidance as they develop increasingly independent research competencies.

Here, we use longitudinal student data across three semesters of a Bioinformatics CURE to examine how students incorporated GenAI into the development of core research competencies. Drawing on sociocultural perspectives, we conceptualize GenAI as a mediational tool that can provide feedback, mediate learning, and support collaborative inquiry. Using student artifacts, reflections, and interviews collected across the research experience, we ask: (1) How does AI support personalization in learning (RQ1)? (2) How does AI influence collaboration in CURE environments (RQ2)? (3) How does AI function as a support system for research competency development (RQ3)? By tracing these interactions over time, we identify consistent themes in how the use of GenAI develops alongside student participation in authentic research. We further identify novel ways in which GenAI provides alternative pathways for learning, collaboration, and research practice among learners. Collectively, our findings provide a learner-centered framework for understanding when and how GenAI can extend the instructional support available within CUREs while preserving the active intellectual engagement required for the development of research competency.

## 2. Theoretical Framework

This study is grounded in Sociocultural theory , which conceptualizes cognition as socially situated and shaped through interaction with cultural tools and other participants (Vygotsky 1978). A central concept is the zone of proximal development (ZPD), which describes the difference between what learners can accomplish independently and what becomes possible with appropriate mediation (Vygotsky 1978). Mediation refers to the temporary and adaptive support within this space, which enables learners to engage in forms of reasoning or problem-solving that exceed their current independent capabilities (Vygotsky 1978; as reviewed in Shabani et al. 2010). Recent work has extended this logic to GenAI, positioning large language models as mediators that can offer cognitive scaffolding when students remain actively engaged with guided interactions (Khan 2024; Liu and Wang 2024; Godoy 2026).

Sociocultural perspectives also emphasize that cognition is distributed across people, tools, and the environments in which activity occurs (Hutchins 1996). This perspective is particularly relevant to CUREs, where research knowledge is developed collaboratively through interactions among students, instructors, and other resources (Auchincloss et al. 2014; Buchanan and Fisher 2022). Within this system, GenAI may reshape how ideas are generated, communicated, evaluated, and acted upon, while also influencing how intellectual labor is distributed across members of a research group. Together, these perspectives provide a framework for examining GenAI as both an individualized scaffold for learning and a mediational resource embedded within collaborative research activity.

## 3. Literature Review

### 3.1 Learner Support and Collaborative Research in CUREs

Research competency in CUREs develops through sustained participation in the uncertain and iterative practices of scientific inquiry (Auchincloss et al. 2014), allowing learners to engage in higher levels of scientific reasoning and critical thinking (Brownell et al. 2015; Corwin et al. 2018). Often, productive learning occurs when research does not proceed as expected. Students encountering substantial experimental challenges can still report gains in confidence, problem solving, and understanding of the research process when they remain engaged in interpreting and responding to those challenges (Gin et al. 2018). However, these outcomes are not guaranteed and depend in part on the quality of guidance students receive as they navigate unfamiliar research practices. Across undergraduate research environments, high-quality mentorship has repeatedly been associated with greater scientific self-efficacy, scientific identity, and development of research performance (Estrada et al. 2018; Monarrez et al. 2020; Goodwin et al. 2022, 2023; Shortlidge et al. 2023). As CUREs commonly operate in many-students-to-one-instructor environments, this creates trade-offs between expanded access and the individualized professional guidance characteristic of apprenticeship research experiences (Burmeister et al. 2021). Recent CURE models have therefore incorporated teaching assistants and peer mentors specifically to expand research guidance, including tailored support for technical tasks, scientific writing, feedback, and increasingly independent research activity (Goodwin et al. 2023; Neely et al. 2025; Panzik et al. 2025). The degree to which GenAI can augment the responsive support that students need as their research evolves is only beginning to be explored (McLaughlin et al. 2025; Weston and Long 2026).

In addition to direct mentorship, working with peers during CUREs provides opportunities to improve scientific reasoning and develop the intellectual and communication practices associated with collaborative science (Auchincloss et al. 2014). In professional research teams, collaboration frequently depends on a division of labor in which members contribute differentiated expertise to complementary components of a shared scientific problem and coordinate specialized work into a coherent research product (van Knippenberg et al. 2004; Zhu and Wholey 2018; Haeussler and Sauermann 2020). CURE teams also require students to coordinate research tasks, exchange information, and contribute jointly to a research objective (Auchincloss et al. 2014; Werth et al. 2022). However, task specialization has different implications when team members are novice researchers whose expertise is still developing. Students commonly allocate tasks according to prior expertise and preference, which can limit opportunities for individuals to practice unfamiliar skills (van der Meulen and Aivaloglou 2021) and shape unequal power dynamics that potentially concentrate decision-making among students perceived as more expert (Salomon and Globerson 1989; Huynh et al. 2020). The extent to which students leverage GenAI tools to alleviate the pedagogical tension between efficiently accomplishing authentic team-based research and ensuring they participate broadly enough to develop research competency remains unknown.

### 3.2 Integration of GenAI tools into CUREs

Recent work increasingly suggests that GenAI can act as a form of scaffolding in which technological support expands what learners can accomplish while preserving their responsibility for interpretation and decision-making. For example, GenAI has been shown to support higher-order thinking, problem-solving, and greater learner autonomy when students remain actively engaged in questioning, evaluating, and revising AI-generated outputs (Bai and Wang 2025; Deng et al. 2025; Sánchez Muñoz et al. 2025). Similar benefits have been observed when GenAI is used to extend formative feedback or support collaborative processes (Noroozi et al. 2025; Wei and Perkins 2026). However, this relationship changes when learners delegate the underlying cognitive work to the technology. Overreliance on GenAI has been associated with cognitive offloading, reduced critical engagement, and weaker independent reasoning when learners rely on generated outputs rather than interrogating or extending them (Bastani et al. 2025; Lee et al. 2025; Choudhuri et al. 2026; Liu et al. 2026). This has led to concerns that GenAI tools may harm long-term cognitive development (Cicchetti 2024; Devaki 2024; Gerlich 2025; El Hage 2026) with recent work suggesting that students with traits classically associated with STEM, such as higher computer and technological expertise, may be particularly vulnerable to overreliance on GenAI (Choudhuri et al. 2026). The distinction between GenAI that supports participation in complex cognitive activity and GenAI that erodes it is particularly consequential for CUREs where the acquisition of independent research decision-making and scientific reasoning skills are themselves target learning outcomes. This raises the question of how students choose to move between these fundamentally distinct forms of AI engagement over the course of an authentic research experience.

## 4. Methods

### 4.1 Study Overview

This study was conducted under approval from the University of XXXX Institutional Review Board (IRB-25-0458) and draws on a longitudinal qualitative dataset collected across three semesters of an honors section of BINF 1101: Introduction to Bioinformatics and Genomics. BINF 1101 is a general education introductory undergraduate course. The honors laboratory section was redesigned as a course-based undergraduate research experience centered on the intersection of GenAI, bioinformatics, and genomics. The CURE was developed collaboratively by the second and third authors with support from internal university funding and was designed to engage students in the complete research process, from question formulation and experimental design through data collection and curation, analysis, interpretation, scientific writing, and dissemination. Rather than completing a sequence of predetermined laboratory exercises, students worked in small research teams to develop original questions about the use or performance of GenAI in bioinformatics and genomics and progressively developed these projects throughout the semester. At the beginning of the course, teams were given the option to select from a curated set of instructor-developed research topics or to formulate a project of their own design, provided that the research question meaningfully intersected with bioinformatics or genomics. This structure was intended to provide an accessible entry point for students who preferred greater guidance while preserving substantial latitude for teams to pursue questions aligned with their own interests. Course learning outcomes consequently emphasized the ability to design a research study, curate and analyze data, communicate findings, work effectively as a research team, and disseminate results through an academic conference-style research presentation.

The instructional sequence followed a gradual release of responsibility model in which responsibility for research decisions shifted progressively from instructors toward student teams as their projects developed (Pearson and Gallagher 1983; Fisher and Frey 2008; Webb et al. 2019). Each week began with an instructor-guided discussion focused on the research practices required for the next stage of the project, followed by student groups advancing their projects independently. Early sessions introduced students to GenAI, prompting, research-question development, and study design. This was followed by moving through literature review, data collection and curation, statistical analysis, and interpretation. Student projects culminated in a scientific presentation to other sections of the course as part of a simulated research conference experience, focusing on scientific communication and dissemination to a general audience. Later sessions increasingly emphasized peer and instructor feedback, presentation development, interpretation of findings, and independent articulation of each project's implications and limitations. This progression was intended to provide sufficient guidance for novice researchers while progressively transferring responsibility for methodological and interpretive decisions to the students.

GenAI was intentionally embedded throughout this research process rather than confined to a single assignment or instructional unit. Students were encouraged to use GenAI during laboratory activities and project development for purposes they considered useful. Importantly, students were not prescribed a required frequency or amount of AI use, nor were they instructed to avoid GenAI for particular stages of the research process. Instead, instructional activities emphasized critical engagement with generated content, encouraging students to examine the strengths, limitations, and usefulness of AI outputs in relation to their own research goals. This design created a learning environment in which all students had explicit permission and repeated opportunities to engage with GenAI while retaining discretion over whether, when, and how they incorporated AI support into their research practice. This design allowed us to examine student choices surrounding GenAI use as an evolving component of research activity rather than as a standardized instructional treatment.

### 4.2 Positionality Statement

The research team occupied complementary positions in relation to the CURE examined in this study. The first author was an undergraduate researcher who had previously completed the course and therefore brought direct experiential knowledge of the instructional environment, research activities, and ways students encountered GenAI during the CURE. The third author was the course instructor in Bioinformatics & Genomics who participated in the design and implementation of the CURE across the study period. The second author, a faculty researcher in education, collaborated in the initial conceptualization and design of the CURE and conducted scripted interviews, but was not responsible for day-to-day course instruction. We treated these differing relationships to the study context as both analytic resources and potential sources of interpretive bias. The perspective stemming from the prior participation of the first author provided sensitivity to features of the student experience that may have been less visible to faculty researchers. In contrast, the instructional role of the senior author provided detailed knowledge of the pedagogical intentions and evolution of the CURE. The second author contributed an external educational perspective that supported interpretation beyond the assumptions embedded in course design and implementation. Throughout analysis, interpretations were developed collaboratively across these perspectives rather than relying on the account of a single researcher. This reflexive structure was intended to make researcher positionality explicit while using differences in proximity to the course as a resource for interrogating emerging interpretations of student learning, collaboration, and GenAI use.

### 4.3 Participants

The present study took place at a large public university on the east coast of the United States, in which this course fulfills the general education requirement of a natural science course with a laboratory section. This study focuses on a single laboratory section of the larger course that fulfills the same general education requirement for students in the honors college. The honors lab met once per week for 120 minutes. Participation in the research study was voluntary and independent of enrollment or performance in BINF 1101. All students completed the same course activities regardless of whether they elected to participate in the research study. Students were invited to opt into the study through an IRB-approved consent process, and participation decisions were not disclosed to course instructors until after completion of the semester and submission of final grades. Students who declined participation or did not submit a consent response were treated as non-participants, and their course materials were excluded from the research dataset. Across the three study semesters, 18 students (5 male & 13 female) consented to participate, representing ~82% of students enrolled in the honors CURE. Participants represented freshmen through upper level students and a range of STEM and non-STEM majors. All study materials were de-identified prior to analysis, and participants were assigned randomly selected pseudonyms for presentation of qualitative data. Potentially identifying details such as unique majors were generalized or omitted where necessary in this study.

### 4.3 Data Collection

We collected three data sources that were designed to capture perspectives on how the relationship between GenAI and the research process developed for each student. The first data source was the video recording  of each team's research presentations in the simulated research symposium. These presentations were conducted in the style of a 20-minute research talk at an academic conference and provided evidence of how students collectively articulated their research questions, partitioned teamwork and methodological decisions, and understood their findings and broader implications of their projects. The second data source was  video-recorded individual post-presentation interviews structured to simulate the scholarly conversation that follows a professional research presentation. Interviews asked students to describe their contributions to the research team, discuss the results and implications of their project, and also explain how GenAI intersected with their project. This included asking how they used GenAI, considering whether their AI practices and beliefs changed during the semester, identifying challenges or conceptual advances associated with AI use, and reflecting on the future role of GenAI in research and learning. Finally, the third data source was a final written assignment in which students were asked to 1) generate each section of a scientific research paper using AI and then critically evaluate the output and 2) to consider how their own understanding and use of AI had changed over the semester, thereby providing an opportunity for retrospective assessment of learning and AI use. Considered together, these sources provided complementary evidence of how students described individualized AI support, negotiated AI within collaborative research, and understood its role in their development as increasingly independent researchers.

### 4.4 Data Analysis

We conducted a qualitative analysis of our data using reflexive thematic analysis (Braun and Clarke 2021), as our aim was to identify and interpret patterns of shared meaning across accounts of GenAI use, collaborative research, and developing research competency rather than to quantify the prevalence of predetermined categories. Analysis proceeded recursively through six phases of reflexive thematic analysis: familiarization with the dataset, coding, generating initial themes, developing and reviewing themes, refining, defining, and naming themes, and producing the analytic account (Braun and Clarke 2021). The first author uploaded transcriptions of all data sources to NVivo 15.3.2. She began by reading the complete qualitative corpus and conducting an initial round of coding. Coding was primarily inductive, allowing patterns to emerge from student accounts, while remaining theoretically informed by sociocultural concepts of mediation and distributed cognition relevant to the study questions. Codes were developed at both semantic and latent levels, capturing the explicit descriptions of student experiences as well as underlying assumptions and patterns of meaning concerning learning, collaboration, independence, and GenAI use. The same coding approach was used for all three research questions.

Theme development was subsequently conducted through an iterative and collaborative process involving all authors. Codes and associated data extracts were examined for broader patterns of shared meaning, and candidate themes were repeatedly compared with both coded extracts and the dataset as a whole. Through successive analytic discussions, candidate themes were reorganized, combined, divided, discarded, or reconceptualized as the team refined the central organizing concept underlying each theme. These discussions were not intended to establish coder agreement or a single objective interpretation of the data. Instead, differences among researchers were used reflexively to interrogate assumptions, consider alternative interpretations, and strengthen the coherence of the developing analytic account. Theme development continued until the final themes provided a coherent account of how students described personalization, collaboration, and research competency development through their engagement with GenAI.

## 5. Findings

We organize the findings around the three research questions, grouping closely related themes into sections on personalization, collaboration, and research competency. We begin with how GenAI supported student-directed inquiry, then examine its role within research teams, and finally consider how students developed critical judgment and revised their conceptions of GenAI. As themes frequently co-occurred within student research experiences, this sequence aligns the presentation of findings with the research questions rather than suggesting that these dimensions developed independently.

### *5.1 GenAI became an on-demand support system across the research process*

Across semesters, a prominent theme was students describing GenAI as a supporting resource that could be adjusted to their knowledge level at a particular point in the research process. For several students, this support took the form of an individualized tutor for disciplinary material that initially exceeded their background knowledge. Caleb described encountering unfamiliar terminology while reading the scientific literature and deliberately changing the level at which AI explained it:

> I spent a lot of time at the beginning of our project kind of taking these big bio terms that I really don't know anything about. I spent an afternoon one day, and I just put them all in, and I said, teach this to me like I'm in third grade. Like, just explain it to me so I can at least get that base understanding to know what I'm looking at when I'm reading these papers.

For Caleb, the distinction between this interaction and more static instructional resources was the ability to continually reshape an explanation until it became accessible. As he later explained, “when you're reading a paper and listening to a lecture, it's kind of given to you one way," whereas with AI he could simply ask, "hey, reword this. so I can understand it." The same process extended to technical skills that would otherwise have represented substantial barriers to participation. Aadhya, a business major with little prior coding experience, described GenAI as making computational research accessible despite entering the CURE without the disciplinary preparation of some peers:

> I feel like it definitely made a lot of things accessible to people. Like, you don't have to be in the STEM major to use AI or, you know, do stuff you can't do without being in STEM. So I feel like it really opens that bridge and lets everyone be a part of it, and like, traditionally, like, coding would have taken a while for me to learn, but AI made it easier for me as a business major to quickly use it efficiently, and to learn from it as well.

Other students described similar uses when encountering unfamiliar computational environments. Siya, for example, explained, "I didn't really know how to use R studio because I'm more familiar with, like Python and Java as a data science major." Therefore, she used ChatGPT to determine how to implement the coding needed for the group's analysis.

A theme common to each semester was students using this individualized support to expand and refine what they could imagine doing with their research prior to analysis. Caleb recalled that his group began project development by discussing their personal interests and then using GenAI to translate those interests into possible research questions: "I said I like to exercise, I like to work out, so we kind of put that into an AI and said, hey, what are some research topics we can look into using AI based on exercise?" Aisha similarly described AI-generated alternatives as expanding the range of possibilities she considered, explaining that "it gives you options, and it, like, makes me think, like, oh, I never thought about it like this," and later summarized this experience by noting, "There's, like, more ways of thinking that you might have not thought of, but, like, the AI sort of helps you."

### *5.2 GenAI empowered student-directed inquiry to expand the conceptual reach of the CURE*

CUREs developed and taught by individual faculty almost universally follow a model in which they are focused around the instructor areas of research expertise (Brownell et al. 2012; Ward et al. 2014; Shortlidge et al. 2016). However, across semesters, none of the research teams selected a topic that overlapped with the disciplinary domains of the instructor which lie in phylogenetics and biodiversity (Dornburg et al. 2011, 2014, 2022; Dornburg and Near 2021) , comparative genomics and immunogenetics (Dornburg and Yoder 2022; Carlson et al. 2023; Mallik et al. 2025; Yoder et al. 2025; Barker et al. 2026), and evolutionary epidemiology (Townsend et al. 2021, 2022, 2023, 2025). Moreover, topics selected only marginally overlapped with the prior focus of the CURE on topics in evolutionary medicine. Instead, every team developed a distinct research question, producing projects that extended across human health, exercise physiology, medical decision-making, forensics, environmental science, and the social consequences of emerging AI technologies (**Table 1**). GenAI was frequently identified as helping students transform existing interests into questions they felt capable of investigating. Kiran described this emerging through the intersection of her interests within the group: "he was really into sports. And I am a dancer myself, so we were like, oh, like, let's do some sort of, like, exercise-related thing, and then I was like, why not, like, epigenetics?" Other teams followed different trajectories. Siya explained, "I'm really into like researching minorities or like disparities between communities in multiple aspects, whether it's like healthcare or any other social construct," which contributed to a project examining demographic bias in AI-supported medical decision-making. Maya similarly connected her topic to previous work in medical AI, explaining that she "had already a little bit of background with this topic" and that her group members "were interested in it as well. So we were able to kind of just run with that."

**Table 1. Research domains and topics selected by student research groups each semester**

| Semester \| Group | Domain | Research question or topic |
|---|---|---|
| Spring 2025 \| Team 1 | Environmental informatics and water quality | Using environmental data to predict variation in water quality with an AI model |
| Spring 2025 \| Team 2 | Sports informatics and performance analytics | Integrating publicly available training, performance, and biosensor data to optimize athletic performance across levels of competition |
| Spring 2025 \| Team 3 | Health informatics and data science | Examining how biases in medical data are reflected in AI models and may contribute to healthcare disparities among minority patients |

| Fall 2025 \| Team 1 | Cancer genomics and personalized medicine | Evaluating AI-generated personalized treatment plans across different cancer types using simulated patient data |
|---|---|---|
| Fall 2025 \| Team 2 | Forensic and machine learning | Using machine learning to improve the analysis of degraded and mixed DNA samples in forensic investigations |
| Spring 2026 \| Team 1 | Genomic medicine and data science | Testing whether AI can identify disease-associated alleles or whether predicted disease risk is influenced by demographic biases using simulated genomic data |
| Spring 2026 \| Team 2 | Health informatics and psychology | Testing whether demographic biases influence AI identification of disease-associated alleles and predicted disease risk using simulated genomic data |
| Spring 2026 \| Team 3 | Epigenetics and exercise physiology | Modeling the effects of exercise on biological aging using epigenetic measures of the biological clock |

The analytical computational skills needed to extend the conceptual reach of projects were largely fueled by student engagement. However, engagement arose from the interaction between student-directed inquiry and AI-mediated support rather than from AI use alone. Maya selected a project closely aligned with her prior interests and stated that GenAI itself did not increase her connection to the work: “I wouldn't say it made me feel more connected to the content of my project. I more so kind of just, would use it as, like, a tool of kind of helping sort through information.” Students also described using GenAI to participate in aspects of their projects that initially fell outside their expertise. Selene, who described herself as “very anti-code since high school,” reflected that she “didn't really realize how much I had learned about it until, like, I was explaining it to my groupmates,” calling the addition of coding to her skills “definitely a big breakthrough.” Aadhya articulated a similar change in how she understood the relationship between assistance and independence. She explained that she initially “used AI to help me find answers without me learning the stuff it gave me,” but that the project later “really made me think deeper on how and why AI should be used as a tool, not as a replacement to our learning.”

Bianca explicitly connected the structure of the CURE to the diversity of projects that emerged, noting, “you kind of gave us freedom to really choose, you know, any topic that we wanted, which then, in turn, like, made us have three… all three unique projects.” This investment manifested into project ownership. Adesh characterized the experience as one in which “we just had a professor looking over us, you know, and like just guiding us. But it felt like we're doing all that. We're doing all the work on our own. And like the creation. And he just was a support.” Kiran similarly distinguished the experience from receiving information passively:

> There was something different about actively interrogating a topic versus possibly receiving the information.” She continued, “since I had to frame the questions and evaluate the answers and kind of decide what to look into next. I felt I was actually being invested in the research in a way that just reading an assigned paper wouldn't always.

For some students, this investment and ownership became visible in their preference for working on the research over other course activities. Amie recalled that when an activity felt removed from the project, “Everyone was kind of like uninvested in it. They're like, well, we just want to work on our project. We don't want to do this like side tangent thing.”

### *5.3 GenAI mediated student collaboration and task differentiation*

Distributing research tasks around individual differences in disciplinary background and existing skills emerged as a nearly ubiquitous theme for each student team during each semester. Aisha, a data science major working with a biology major, described this division directly:

> I looked more at the data aspect of it, and, like, the AI aspect of it, because I… my major is data science, and I guess my, like, partner, she looked more at, like, the biological aspect of it, because she was a biology major. So, like, kind of just kind of fit into our majors and stuff like that.

A similar pattern emerged in a team examining AI and athletic performance. Adesh explained:

> Yeah. So I mean, we all were like general, like creation and stuff, you know, like, I mean, putting together the pieces. But once data collection started I was building the code. So I was basically like the software developer for the team. Chris was the researcher. And Brooklyn was the like data collector.

These divisions resembled interdisciplinary research teams in which members contributed differentiated forms of expertise toward a shared problem. However, specialization did not isolate students within independent pieces of the project. Instead, students reported using GenAI in realtime to bridge disciplinary gaps between team members. Jordan described using AI specifically to follow conversations occurring within the group:

> it did help me feel more connected in the sense that because I could ask it questions. of, like, what exactly the other people in my group were talking about. It was able to… like, I was able to, like, ask follow-up questions just to, like, solidify my understanding.

GenAI also allowed students with different levels of prior expertise to engage with the same technical component of a project in different ways. Adesh contrasted his own use of AI as an experienced programmer with that of a biology-major teammate who had no prior coding experience:

> He literally said in the presentation he has 0 experience, you know, and he just put all the stuff in all the research and everything into ChatGPT. And it generated all the code for him, whereas me like I have experience. So I was using AI to just help me summarize my data and everything, you know.

This contrast reflected a general phenomenon in which students reported using GenAI to make boundaries between roles more permeable and contribute to tasks in the subject areas of other group members. Kiran, for example, reported, "I did a lot of the background information," while also contributing to statistical interpretation and helping a data science teammate decide how analytical categories should be structured. Selene similarly described her contributions as concentrated first in background research and later in data formation and visualization.

In a few instances, teams collectively lacked expertise required by their project. Here they used GenAI as shared research infrastructure, allowing their team of novices to move through a technical task for which no member initially occupied a clear expert role. Aadhya recalled that "my groupmates and I, we didn't have much experience using R [a programming language], so we asked ChatGPT, like, what kind of code we might need to type and how to use

R, and it really helped us.” The group was subsequently able to generate the visualizations required for its analysis.

### *5.4 GenAI expanded research capability, but competency required critical judgment*

Across semesters, students frequently described GenAI as expanding the range of research activities they felt capable of undertaking, but noted that it required critical evaluation. Haruka reflected, “Prior to this, I had not considered AI as a resource for coding, and it significantly expanded what I felt capable of doing as an undergraduate researcher.” For Haruka, this expanded capability did not translate into unquestioned trust in the technology. Reflecting on the same experience, she explained,

> I didn't know that it could write code, I didn't know that it could do all of these things, but it also showed me the limitations through this project as well. And learning to be critical of the output, and not always just taking what it says as face value.

This ability to evaluate AI outputs emerged as a dominant theme. Adesh recognized how effective use of GenAI required his own disciplinary understanding: “I just have to be knowledgeable enough about it to check it and make sure that what I'm generating is actually useful and actually like relevant to what I'm doing.” Kiran articulated a similar shift more explicitly: “If I determined that I could not critically analyze an output received from the AI tool regarding its accuracy, that indicated to me that I needed to review the original literature again.” She described developing an iterative process of “prompt, assess, validate, revise” that was “much more methodical than I anticipated and greatly enhanced my comprehension of epigenetic concepts.”

Students also challenged AI outputs with analyses of their own. Haruka described testing a pattern that ChatGPT had failed to identify clearly in her group’s dataset: “I was able to create a logistic regression in RStudio that proved that our dataset was correct and the correlation did exist.” She found it particularly notable that “ChatGPT was not able to quickly identify … what we thought was a pretty clear pattern in the data.” For several students, this changing relationship with GenAI also altered how they understood scientific practice. Bianca summarized this shift by explaining,

> This course showed me that being a scientist is not just about using advanced tools, but about questioning results, testing assumptions, and making evidence-based conclusions. That shift in mindset is probably the most valuable thing I have gained this semester.

**5.5 Conceptions of GenAI shifted from shortcut to conditional research partner**

Students entered the CURE with markedly different expectations about the legitimacy and usefulness of GenAI in academic work that were shaped by their perceptions of other courses where AI was associated primarily with cheating, unreliable information, or threats to academic practice. Amie reflected, "I was incredibly skeptical of it at first, as for a while I have only heard of it as a method of cheating or faking essay writing." Haruka similarly described her starting position more strongly, writing that she "viewed it more as a potential threat to rigorous science than as a useful tool." Aisha similarly explained that she initially saw AI "as a convenience but not a tool for real academic work" and assumed that research use was either "cheating" or unlikely to produce trustworthy results.

Over each semester, there was a marked change in perspective after using AI as indicated by Aisha who noted that it "can be a powerful research partner when used responsibly." Amie similarly described discovering uses she had not previously considered, including "using it as an editing tool, an assistant, a peer, or having it 'pretend' to be someone to help you with something." Kiran noted that her "perception of AI's role in conducting scientific research evolved from viewing AI as a means of shortcutting into perceiving AI as an accelerator." Adesh drew an even sharper boundary, arguing that researchers should "recognize AI as a collaborator, not a shortcut, not a leader."

By the end of each semester, students commonly noted that these shifts placed them at odds with broader academic norms surrounding AI. Aisha noted that "using AI in school right now" carries "a very negative connotation" and can be interpreted as "cheating," whereas having explicit permission to use AI throughout a semester-long research project showed her that "you can still use AI and learn from it." Kiran similarly contrasted the course with what she perceived as "a huge anti-AI sentiment" in biology and public health, adding that projects such as this "open your eyes to, like how useful AI really can be, and if you use it correctly, it's not the end of the world."

## 6. Discussion

This study examined how GenAI shaped personalization, collaboration, and research competency within an authentic undergraduate research environment. Across three semesters, GenAI consistently expanded the range of research activity that novice students could enter by providing individualized, on-demand support and helping teams coordinate across differences in disciplinary expertise. This support allowed students to pursue questions well beyond the research domains of the instructor, distribute work among teammates according to emerging expertise, and engage with conceptual and technical tasks that would otherwise have been difficult to access. However, these benefits were not automatic. Productive use depended on students progressively developing the disciplinary knowledge and critical judgment needed to selectively incorporate AI-generated contributions. Human mentorship and peer interaction remained central to framing questions, interpreting evidence, and coordinating research decisions. Taken together, our findings suggest that GenAI can extend the reach of CUREs not by replacing traditional elements such as mentorship, disciplinary reasoning, or collaborative expertise, but by becoming an additional mediational resource through which students assume greater responsibility and research independence.

### *6.1 GenAI Expanded the Intellectual Reach of the CURE*

CUREs are typically built around areas in which instructors can provide substantial disciplinary and methodological expertise, allowing students to engage in authentic inquiry while receiving the guidance needed to navigate unfamiliar research practices (Shortlidge et al. 2016; Soto et al. 2025). For novice researchers, this mentorship is particularly important because authentic inquiry routinely introduces conceptual and methodological problems that exceed their existing expertise (Shanahan et al. 2015; Goodwin et al. 2023). However, our findings suggested that GenAI can relax the dependence of student-directed inquiry on the boundaries of instructor expertise. All research teams used GenAI to expand the curriculum outward toward problems they themselves found meaningful that generated distinct topics well beyond the research expertise of the instructor (**Table 1**). We observed that GenAI served to provide mediation, creating a  zone of proximal development calibrated to the needs of a student (Vygotsky 1978). Students turned to GenAI for mediation of problems that exceeded their immediate capabilities such as adjusting explanations of unfamiliar concepts or using GenAI to cross substantial technical barriers. This mediation was initiated by students in response to problems arising from their own projects, meaning that it did not depend on the instructor anticipating every conceptual or technical barrier in advance. By sharing the responsibility of mediation, GenAI allowed the instructor to respond to emerging questions, provide feedback, support methodological and interpretive decisions, and in some cases learn new material alongside students. This outcome extended observations that AI-supported learning can provide responsive and individualized guidance (Yilmaz and Karaoglan Yilmaz 2023; Bai et al. 2025; Kestin et al. 2025).

Undergraduate course based research experiences generally follow a model in which the intellectual autonomy of the learner is restricted to the overarching research domain established by the instructor or an existing research program (Auchincloss et al. 2014; Shortlidge et al. 2016; Soto et al. 2025). However, our results suggest that GenAI can allow mentorship in CURE settings to more closely resemble mentorship in independent graduate research environments that typically provide a far greater degree of intellectual freedom (Minshew et al. 2021; Nuis et al. 2023; Wang et al. 2026b). Students explicitly linked this freedom to greater investment and ownership, describing how active interrogation of personally meaningful questions differed from engaging with material selected for them. This extends observations that GenAI-generated content aligned with the individual interests of learners can increase motivation, interest, and engagement (Bai et al. 2025; Tasdelen and Bodemer 2025) and suggests that GenAI can expand the conceptual reach of the research experience by helping transform student interest into researchable inquiry. In turn, this can create a learner-directed form of personalization in which students themselves determine the intellectual territory a course explores.

### *6.2 GenAI bridged novice collaboration and interdisciplinary team science*

Professional interdisciplinary research teams commonly coordinate complex scientific problems by distributing work across members with differentiated expertise (Lewis 2003; Ren and Argote 2011; Haeussler and Sauermann 2020). For teams comprised entirely of novice undergraduate researchers, independently achieving this level of coordination is considerably more difficult (Karsai et al. 2011; Cheruvelil et al. 2020). However, our findings suggested that GenAI allowed students to approximate the distributed expertise characteristic of interdisciplinary science while they were still developing the expertise needed to participate in it. Students consistently divided work according to their disciplinary backgrounds and existing skills, while GenAI helped them communicate across those divisions by explaining unfamiliar concepts, supporting follow-up questions about the contributions of teammates, and providing technical assistance when expertise was absent from the team. In this sense, GenAI became part of the distributed cognitive system of the team rather than replacing the expertise of individual members, extending emerging observations that GenAI can facilitate collaborative knowledge construction and exchange (An et al. 2025; Wei and Perkins 2026). Importantly, this organization did not resolve a well-known pedagogical tension between functioning effectively as a research team and ensuring that every student practices every research skill (van der Meulen and Aivaloglou 2021). In some cases, students used GenAI to enter unfamiliar areas, learn new technical skills, or assist peers with discipline-specific tasks, whereas in others students remained primarily within their established roles. If the instructional goal is for every learner to practice each component of the research process, GenAI alone is therefore unlikely to produce that outcome consistently. Deliberate instructor-mediated structures such as task rotation, cross-training, or shared responsibility would still be required when broad individual skill development is itself a learning objective.

### *6.3 Research independence required judgment, not the absence of AI*

Concern that GenAI may enable students to complete academic tasks without engaging in the cognitive processes those tasks are intended to develop has prompted many instructors to restrict its use, redesign assessments, or explicitly regulate when AI assistance is permissible (Ali et al. 2024), even when broader institutional policies remain permissive or encourage responsible integration (Wang et al. 2024). These concerns are supported by growing evidence that uncritical reliance on GenAI can promote cognitive offloading and weaken independent reasoning, creating situations in which AI improves immediate task performance without producing equivalent gains in learning or subsequent unaided performance (Bastani et al. 2025; Jose et al. 2025; Tian and Zhang 2025). However, our longitudinal findings suggested that AI-supported capability and research competency both increased as students developed the disciplinary knowledge needed to evaluate what GenAI produced. Some students explicitly described beginning the semester by using AI to obtain answers, whereas later use involved checking outputs against primary literature, testing AI-generated interpretations with independent statistical analyses, and recognizing when they lacked sufficient knowledge to determine whether an output was trustworthy. This pattern is consistent with emerging evidence that the educational consequences of GenAI depend substantially on whether learners critically evaluate and regulate its use rather than treating generated responses as authoritative (Shi et al. 2025; Yan et al. 2025). From a sociocultural perspective, our findings suggest that research independence can be obtained through intellectual responsibility with GenAI, in which learners retain authority over what questions are asked, whether evidence is accepted or external validation is required, and if an AI-generated contribution is ultimately incorporated into the research process or not.

As GenAI becomes increasingly embedded in education and professional practice, AI literacy is highly correlated with productive academic use of GenAI (Tzirides et al. 2024; Xiao et al. 2024; Singh et al. 2025). Rather than treating AI literacy as something that must be established before students encounter GenAI, our results suggest that authentic disciplinary use can itself provide the conditions through which such literacy develops. Participants generally entered each semester with views of GenAI as primarily a cheating tool, a threat to rigorous academic work, or merely a convenient source of answers. Through sustained use and experimentation, students increasingly described GenAI conditionally as an assistant, collaborator, peer, or accelerator while simultaneously articulating boundaries around those roles that placed themselves in positions of authority. These results are consistent with observations that learners applying AI to authentic tasks and reflecting on the quality and consequences of its outputs can shift their attitudes and promote more differentiated forms of AI literacy (Tzirides et al. 2024; Suh 2025). Such a heightened awareness surrounding the utility and limitations of GenAI tools may be particularly consequential for students entering scientific research. Failures to critically verify AI-generated claims, references, or analyses are increasingly implicated in threats to research integrity and the accumulation of low-quality AI-assisted scholarship (Giray et al. 2026; Resnik and Hosseini 2026; Thorp 2026). Preparing students to use GenAI critically within authentic research may therefore be far more urgent than filling an instructional response to a new technology and could be vital for preparing early career researchers for career skills capable of mitigating a looming crisis in scholarship.

### *6.4 Instructional implications and boundary conditions.*

The benefits observed in this CURE should not be interpreted as evidence that unrestricted access to GenAI is sufficient to promote learning. Students were given broad discretion over when and how they used GenAI, but this freedom operated within a highly structured research environment that repeatedly required them to explain methodological choices, evaluate generated outputs, justify interpretations, reflect on limitations, and articulate the reasoning underlying their decisions. GenAI was incorporated throughout the assignments themselves, with students routinely asked not only to use the technology but also to examine and reflect on its contributions throughout a scaffolded research process that spanned question development to research dissemination. Instructor and teaching-assistant support also remained continuously available as students encountered conceptual and technical challenges. The broader implication is therefore not that all courses should adopt permissive AI policies or redesign instruction around GenAI. Rather, for courses that choose to permit substantial AI use and seek to leverage it for more ambitious forms of student inquiry, our findings suggest that access is most productively paired with pedagogical structures that preserve learner responsibility for explanation, validation, reflection, and decision-making. Although broad access to GenAI may expand what students are capable of attempting, the surrounding instructional design remains critical for ensuring that this expanded capability contributes to actual learning.

***6.5 Limitations and future directions.***

Several limitations constrain the generalizability and causal interpretation of these findings. This study examined a relatively small, self-selected sample from an honors CURE at a single institution in which students were permitted to determine when, how, and how extensively they used GenAI. As such, we cannot isolate which specific patterns of AI engagement were responsible for particular learning outcomes or determine whether similar effects would emerge under more tightly controlled conditions. However, this naturalistic variation also provided a valuable opportunity to examine how students incorporated GenAI when given substantial agency within an authentic research environment. The resulting themes were highly persistent across three semesters. This persistence is especially striking given the substantial variation in research topic, disciplinary background, prior technical experience, and the specific ways individual teams integrated GenAI into their projects. Across heterogeneous teams investigating markedly different questions, students repeatedly described similar patterns of adaptive support, distributed expertise, critical evaluation, and changing conceptions of AI. This consistency strengthens our confidence that the themes capture recurring features of how novice researchers incorporated GenAI within this CURE. Future work should experimentally compare different forms of AI scaffolding, examine how team-level access influences participation in unfamiliar research tasks, and follow students into subsequent research experiences to determine whether gains in critical judgment and research independence persist beyond the course.

An additional challenge is that the educational baseline of future undergraduates is rapidly changing. The students examined here completed most of their K–12 education before widespread access to generative AI and therefore developed many foundational academic practices in environments where GenAI was not routinely available. However, future cohorts will increasingly enter higher education after years of exposure to these tools across educational settings that differ substantially in instructional AI training and opportunities to develop AI literacy (Davin et al. 2026; Deng et al. 2026; Wang et al. 2026a). As a result, students may arrive with very different relationships to GenAI, ranging from extensive experience critically evaluating and strategically using AI to habitual reliance on it for cognitive offloading or task completion. Whether the patterns observed here will persist across this changing developmental landscape is therefore an urgent question. Longitudinal research spanning secondary and postsecondary education will be necessary to determine how accumulated AI exposure shapes these capacities and whether instructional approaches that promote productive AI-supported learning must change as increasingly AI-experienced cohorts enter higher education.

## *7. References*

Abdallah N., Katmah R., Khalaf K., Jelinek H.F. 2025. Systematic review of ChatGPT in higher education: Navigating impact on learning, wellbeing, and collaboration. Soc. Sci. Humanit. Open. 12:101866.

Ali A., Collier A.H., Dewan U., McDonald N., Johri A. 2024. Analysis of Generative AI policies in computing course syllabi. arXiv [cs.CY].

An S., Zhang S., Guo T., Lu S., Zhang W., Cai Z. 2025. Impacts of generative AI on student teachers' task performance and collaborative knowledge construction process in mind mapping-based collaborative environment. Comput. Educ. 227:105227.

Auchincloss L.C., Laursen S.L., Branchaw J.L., Eagan K., Graham M., Hanauer D.I., Lawrie G., McLinn C.M., Pelaez N., Rowland S., Towns M., Trautmann N.M., Varma-Nelson P., Weston T.J., Dolan E.L. 2014. Assessment of course-based undergraduate research experiences: a meeting report. CBE Life Sci. Educ. 13:29–40.

Bai H., Lui W.C., Khiatani P.V. 2025. Promoting student engagement with GPTutor: An intelligent tutoring system powered by generative AI. Int. J. Educ. Technol. High. Educ. 22.

Bai Y., Wang S. 2025. Impact of generative AI interaction and output quality on university students' learning outcomes: a technology-mediated and motivation-driven approach. Sci Rep. 15:24054.

Bangera G., Brownell S.E. 2014. Course-based undergraduate research experiences can make scientific research more inclusive. CBE Life Sci. Educ. 13:602–606.

Barker A.V., Carlson K.B., Wcisel D.J., Birchler De Allende I., Braasch I., Fisk M., Dornburg A., Yoder J.A. 2026. Holostean genomes reveal evolutionary novelty in the vertebrate immunoproteasome that have implications for MHCI function. Mol. Biol. Evol. 43.

Bastani H., Bastani O., Sungu A., Ge H., Kabakcı Ö., Mariman R. 2025. Generative AI without guardrails can harm learning: Evidence from high school mathematics. Proc. Natl. Acad. Sci. U. S. A. 122:e2422633122.

Bell J.K., Eckdahl T.T., Hecht D.A., Killion P.J., Latzer J., Mans T.L., Provost J.J., Rakus J.F., Siebrasse E.A., Ellis Bell J. 2017. CUREs in biochemistry-where we are and where we should go. Biochem. Mol. Biol. Educ. 45:7–12.

Braun V., Clarke V. 2021. Thematic analysis. London, England: SAGE Publications.

Brownell S.E., Hekmat-Scafe D.S., Singla V., Chandler Seawell P., Conklin Imam J.F., Eddy S.L., Stearns T., Cyert M.S. 2015. A high-enrollment course-based undergraduate research experience improves student conceptions of scientific thinking and ability to interpret data. CBE Life Sci. Educ. 14:14:ar21.

Brownell S.E., Kloser M.J., Fukami T., Shavelson R. 2012. Undergraduate biology lab courses: Comparing impact traditionally based "cookbook" authentic research-based courses student lab experiences Journal College Science Teaching. J. Coll. Sci. Teach. 41:36–45.

Buchanan A.J., Fisher G.R. 2022. Current status and implementation of science practices in course-based undergraduate research experiences (CUREs): A systematic literature review. CBE Life Sci. Educ. 21:ar83.

Burmeister A.R., Dickinson K., Graham M.J. 2021. Bridging trade-offs between traditional and course-based undergraduate research experiences by building student communication skills, identity, and interest. J. Microbiol. Biol. Educ. 22.

Carlson K.B., Nguyen C., Wcisel D.J., Yoder J.A., Dornburg A. 2023. Ancient fish lineages illuminate toll-like receptor diversification in early vertebrate evolution. Immunogenetics. 75:465–478.

Cheruvelil K.S., De Palma-Dow A., Smith K.A. 2020. Strategies to promote effective student research teams in undergraduate biology labs. Am. Biol. Teach. 82:18–27.

Choudhuri R., Sanchez C., Burnett M., Sarma A. 2026. Thinking less, trusting more: GenAI's impacts on students' cognitive habits. arXiv [cs.HC].

Cicchetti C.C. 2024. AI in higher education: Does not help, might hurt. J. Educ. Bus. 99:438–443.

Cooper A.C., Delcy M.D., Dolan E.L. 2026. What student struggles do instructors see? Teacher knowledge of students in course-based undergraduate research experiences. CBE Life Sci. Educ. 25:ar7.

Cooper K.M., Blattman J.N., Hendrix T., Brownell S.E. 2019. The impact of broadly relevant novel discoveries on student project ownership in a traditional lab course turned CURE. CBE Life Sci. Educ. 18:ar57.

Corwin L.A., Runyon C.R., Ghanem E., Sandy M., Clark G., Palmer G.C., Reichler S., Rodenbusch S.E., Dolan E.L. 2018. Effects of discovery, iteration, and collaboration in laboratory courses on undergraduates' research career intentions fully mediated by student ownership. CBE Life Sci. Educ. 17:ar20.

Craig P.A. 2017. A survey on faculty perspectives on the transition to a biochemistry course-based undergraduate research experience laboratory. Biochem. Mol. Biol. Educ. 45:426–436.

Daniel K., Msambwa M.M., Wen Z. 2025. Can generative AI revolutionise academic skills development in higher education? A systematic literature review. Eur. J. Educ. 60.

Davin K.J., Kissau S., Haudeck H., Ade-Thurow B., Gómez-Pereira D., Oon P.T., Dornburg A. 2026. Global and educational disparities in AI integration: A study of L2 teacher training and usage patterns. System. 136:103901.

DeChenne-Peters S.E., Scheuermann N.L. 2022. Faculty experiences during the implementation of an introductory biology course-based undergraduate research experience (CURE). CBE Life Sci. Educ. 21:ar70.

Deng R., Jiang M., Yu X., Lu Y., Liu S. 2025. Does ChatGPT enhance student learning? A systematic review and meta-analysis of experimental studies. Comput. Educ. 227:105224.

Deng Y., Çelik F., Duran V. 2026. Governing the unseen: A systematic review of AI literacy among language teachers in higher education. Computers and Education: Artificial Intelligence.:100658.

Devaki V. 2024. Academic integrity and human cognitive development of learners. Advances in Educational Marketing, Administration, and Leadership. IGI Global. p. 195–222.

Dornburg A., Near T.J. 2021. The emerging phylogenetic perspective on the evolution of actinopterygian fishes. Annu. Rev. Ecol. Evol. Syst. 52:427–452.

Dornburg A., Sidlauskas B., Santini F., Sorenson L., Near T.J., Alfaro M.E. 2011. The influence

of an innovative locomotor strategy on the phenotypic diversification of triggerfish (family: Balistidae). Evolution. 65:1912–1926.

Dornburg A., Townsend J.P., Friedman M., Near T.J. 2014. Phylogenetic informativeness reconciles ray-finned fish molecular divergence times. BMC Evol. Biol. 14:169.

Dornburg A., Yoder J.A. 2022. On the relationship between extant innate immune receptors and the evolutionary origins of jawed vertebrate adaptive immunity. Immunogenetics. 74:111–128.

Dornburg A., Zapfe K., Williams R., Alfaro M., Morris R., Adachi H., Flores J., Santini F., Near T., Frédérich B. 2022. Considering decoupled phenotypic diversification between ontogenetic phases in macroevolution: An example using Triggerfishes (Balistidae). bioRxiv.

El Hage S. 2026. Minds growing under algorithmic influence <br> the cognitive price of growing up with AI and what it will cost our future. .

Estrada M., Hernandez P.R., Schultz P.W. 2018. A longitudinal study of how quality mentorship and research experience integrate underrepresented minorities into STEM careers. CBE Life Sci. Educ. 17:ar9.

Fisher D., Frey N. 2008. Better learning through structured teaching: A framework for the gradual release of responsibility. Alexandria, VA: Association for Supervision & Curriculum Development.

Gerlich M. 2025. AI tools in society: Impacts on cognitive offloading and the future of critical thinking. Societies (Basel). 15:6.

Gin L.E., Rowland A.A., Steinwand B., Bruno J., Corwin L.A. 2018. Students who fail to achieve predefined research goals may still experience many positive outcomes as a result of CURE participation. CBE Life Sci. Educ. 17:ar57.

Giray L., Sevnarayan K., Maphoto K.B., Wider W. 2026. AI slop in academic publishing: History, characteristics, manifestations, causes, and mitigation strategies. Internet Ref. Serv. Q. 30:221–244.

Godoy D. 2026. AI mediated scaffolding for second language academic writing in EFL composition classrooms. Discov. Educ. 5.

Goodwin E.C., Cary J.R., Phan V.D., Therrien H., Shortlidge E.E. 2023. Graduate teaching assistants impact student motivation and engagement in course‐based undergraduate research experiences. J. Res. Sci. Teach. 60:1967–1997.

Goodwin E.C., Cary J.R., Shortlidge E.E. 2022. Not the same CURE: Student experiences in course-based undergraduate research experiences vary by graduate teaching assistant. PLoS One. 17:e0275313.

Habib S., Vogel T., Anli X., Thorne E. 2024. How does generative artificial intelligence impact student creativity? Journal of Creativity. 34:100072.

Haeussler C., Sauermann H. 2020. Division of labor in collaborative knowledge production: The role of team size and interdisciplinarity. Res. Policy. 49:103987.

Heim A.B., Holt E.A. 2019. Benefits and challenges of instructing introductory biology course-based undergraduate research experiences (CUREs) as perceived by graduate teaching assistants. CBE Life Sci. Educ. 18:ar43.

Huang A.Y.Q., Chang J.W., Yang A.C.M., Ogata H., Li S.T., Yen R.X., Yang S.J.H. 2023. Personalized intervention based on the early prediction of at-risk students to improve their learning performance. J. Educ. Techno. Soc. 26:69–89.

Hutchins E. 1996. Cognition in the Wild. Cambridge, MA: Bradford Books.

Huynh T., Jambuge A.P., Khong H., Laverty J.T., Sayre E.C. 2020. Positioning in groups: The roles of expertise and being in charge. arXiv [physics.ed-ph].

Johnson D.M., Doss W., Estepp C.M. 2024. Using ChatGPT with novice Arduino programmers: Effects on performance, interest, self-efficacy, and programming ability. J. Res. Tech. Careers. 8:1.

Jose B., Cherian J., Verghis A.M., Varghise S.M., S M., Joseph S. 2025. The cognitive paradox of AI in education: between enhancement and erosion. Front. Psychol. 16:1550621.

Karsai I., Knisley J., Knisley D., Yampolsky L., Godbole A. 2011. Mentoring interdisciplinary undergraduate students via a team effort. CBE Life Sci. Educ. 10:250–258.

Kern A.M., Olimpo J.T. 2023. SMART CUREs: A professional development program for advancing teaching assistant preparedness to facilitate course-based undergraduate research experiences. J. Microbiol. Biol. Educ. 24:e00137–22.

Kestin G., Miller K., Klales A., Milbourne T., Ponti G. 2025. AI tutoring outperforms in-class active learning: an RCT introducing a novel research-based design in an authentic educational setting. Sci. Rep. 15:17458.

Khan W.M. 2024. Analyzing the AI tools'impact on critical thinking in bs English students at Pakistani universities. Journal of Applied Linguistics and TESOL (JALT). 7:1232–1238.

van Knippenberg D., De Dreu C.K.W., Homan A.C. 2004. Work group diversity and group performance: an integrative model and research agenda. J. Appl. Psychol. 89:1008–1022.

Lee H.-P. (hank), Sarkar A., Tankelevitch L., Drosos I., Rintel S., Banks R., Wilson N. 2025. The impact of generative AI on critical thinking: Self-reported reductions in cognitive effort and confidence effects from a survey of knowledge workers. Proceedings of the 2025 CHI Conference on Human Factors in Computing Systems.:1–22.

Lewis K. 2003. Measuring transactive memory systems in the field: scale development and validation. J. Appl. Psychol. 88:587–604.

Liu G., Christian B., Dumbalska T., Bakker M.A., Dubey R. 2026. AI assistance reduces persistence and hurts independent performance. arXiv [cs.AI].

Liu W., Wang Y. 2024. The effects of using AI tools on critical thinking in English literature classes among EFL learners: An intervention study. Eur. J. Educ. 59.

Lopatto D., Hauser C., Jones C.J., Paetkau D., Chandrasekaran V., Dunbar D., MacKinnon C., Stamm J., Alvarez C., Barnard D., Bedard J.E.J., Bednarski A.E., Bhalla S., Braverman J.M., Burg M., Chung H.-M., DeJong R.J., DiAngelo J.R., Du C., Eckdahl T.T., Emerson J., Frary A., Frohlich D., Goodman A.L., Gosser Y., Govind S., Haberman A., Hark A.T., Hoogewerf A., Johnson D., Kadlec L., Kaehler M., Key S.C.S., Kokan N.P., Kopp O.R., Kuleck G.A., Lopilato J., Martinez-Cruzado J.C., McNeil G., Mel S., Nagengast A., Overvoorde P.J., Parrish S., Preuss M.L., Reed L.D., Regisford E.G., Revie D., Robic S., Roecklien-Canfield J.A., Rosenwald A.G., Rubin M.R., Saville K., Schroeder S., Sharif K.A., Shaw M., Skuse G., Smith C.D., Smith M., Smith S.T., Spana E.P., Spratt M., Sreenivasan A., Thompson J.S., Wawersik M., Wolyniak M.J., Youngblom J., Zhou L., Buhler J., Mardis E., Leung W., Shaffer C.D., Threlfall J., Elgin S.C.R. 2014. A central support system can facilitate implementation and sustainability of a Classroom-based Undergraduate Research Experience (CURE) in Genomics. CBE Life Sci. Educ. 13:711–723.

Mallik R., Wcisel D.J., Near T.J., Yoder J.A., Dornburg A. 2025. Investigating the impact of whole-genome duplication on transposable element evolution in teleost fishes. Genome Biol. Evol. 17:evae272.

McLaughlin J.E., Ponte C.D., Lyons K. 2025. Student perceptions of GenAI as a virtual tutor to support collaborative research training for health professionals. BMC Med. Educ. 25:895.

van der Meulen A., Aivaloglou E. 2021. Who does what? Work division and allocation strategies of computer science student teams. 2021 IEEE/ACM 43rd International Conference on Software Engineering: Software Engineering Education and Training (ICSE-SEET).

Minshew L.M., Olsen A.A., McLaughlin J.E. 2021. Cognitive apprenticeship in STEM graduate education: A qualitative review of the literature. AERA Open. 7:233285842110520.

Monarrez A., Morales D., Echegoyen L.E., Seira D., Wagler A.E. 2020. The moderating effect of faculty mentorship on undergraduate students' summer research outcomes. CBE Life Sci. Educ. 19:ar56.

Neely S.H., Meyer Nunes R., Garcia A., Donate C., Benabentos R., McCartney M., Siltberg-Liberles J. 2025. Development of the PeeR Investigators Mentoring Experiences in Research (PRIMER) program: a peer mentoring initiative to increase mentoring in CUREs. J. Microbiol. Biol. Educ. 26:e0004825.

Noroozi O., Haddadian G., Gao X., Schunn C., Alqassab M., Banihashem S.K. 2025. The value

of GenAI for peer feedback provision: student perceptions and impacts. Int. J. Educ. Technol. High. Educ. 22.

Nuis W., Segers M., Beausaert S. 2023. Conceptualizing mentoring in higher education: A systematic literature review. Educ. Res. Rev. 41:100565.

Pallant J.L., Blijlevens J., Campbell A., Jopp R. 2026. Mastering knowledge: the impact of generative AI on student learning outcomes. Stud. High. Educ. 51:714–735.

Panzik J.E., Flom T., Pinnavaia V.M., Natoli A.M., Notley S.F., Light C.J. 2025. Unlocking potential: mentorship training perspectives from undergraduate peer mentors in course-based undergraduate research experiences. J. Microbiol. Biol. Educ. 26:e0000725.

Pearson P.D., Gallagher M.C. 1983. The instruction of reading comprehension. Contemp. Educ. Psychol. 8:317–344.

Poulidis S., Bastani H., Bastani O. 2025. Self-regulated AI use hinders long-term learning. .

Qian Y. 2025. Pedagogical applications of generative AI in higher education: A systematic review of the field. TechTrends. 69:1105–1120.

Ren Y., Argote L. 2011. Transactive memory systems 1985–2010: An integrative framework of key dimensions, antecedents, and consequences. Acad. Manag. Ann. 5:189–229.

Resnik D.B., Hosseini M. 2026. Hallucinated citations produced by generative artificial intelligence may constitute research misconduct when citations function as data in scholarly papers. Account. Res.:2645390.

Rodenbusch S.E., Hernandez P.R., Simmons S.L., Dolan E.L. 2016. Early engagement in course-based research increases graduation rates and completion of science, engineering, and mathematics degrees. CBE Life Sci. Educ. 15:ar20.

Salomon G., Globerson T. 1989. When teams do not function the way they ought to. Int. J. Educ. Res. 13:89–99.

Sánchez Muñoz J.A., Flores-Eraña G., Silva-Campos J.M., Chavira-Quintero R., Olais-Govea J.M. 2025. GenAI as a cognitive mediator: a critical-constructivist inquiry into computational thinking in pre-university education. Front. Educ. 10.

Santillan K.A., Rediske A.M., Olimpo J.T. 2024. Graduate teaching assistants' beliefs and practices regarding mentoring in the context of an online introductory biology CURE: an exploratory study. J. Microbiol. Biol. Educ. 25:e0015024.

Shabani K., Khatib M., Ebadi S. 2010. Vygotsky's zone of proximal development: Instructional implications and teachers' professional development. Engl. Lang. Teach. 3.

Shaffer C.D., Alvarez C., Bailey C., Barnard D., Bhalla S., Chandrasekaran C., Chandrasekaran V., Chung H.-M., Dorer D.R., Du C., Eckdahl T.T., Poet J.L., Frohlich D., Goodman A.L.,

Gosser Y., Hauser C., Hoopes L.L.M., Johnson D., Jones C.J., Kaehler M., Kokan N., Kopp O.R., Kuleck G.A., McNeil G., Moss R., Myka J.L., Nagengast A., Morris R., Overvoorde P.J., Shoop E., Parrish S., Reed K., Regisford E.G., Revie D., Rosenwald A.G., Saville K., Schroeder S., Shaw M., Skuse G., Smith C., Smith M., Spana E.P., Spratt M., Stamm J., Thompson J.S., Wawersik M., Wilson B.A., Youngblom J., Leung W., Buhler J., Mardis E.R., Lopatto D., Elgin S.C.R. 2010. The genomics education partnership: successful integration of research into laboratory classes at a diverse group of undergraduate institutions. CBE Life Sci. Educ. 9:55–69.

Shanahan J.O., Ackley-Holbrook E., Hall E., Stewart K., Walkington H. 2015. Ten salient practices of undergraduate research mentors: A review of the literature. Mentor. Tutoring. 23:359–376.

Shi J., Liu W., Hu K. 2025. Exploring how AI literacy and self-regulated learning relate to student writing performance and well-being in generative AI-supported higher education. Behav. Sci. (Basel). 15:705.

Shortlidge E.E., Bangera G., Brownell S.E. 2016. Faculty perspectives on developing and teaching course-based undergraduate research experiences. Bioscience. 66:54–62.

Shortlidge E.E., Kern A.M., Goodwin E.C., Olimpo J.T. 2023. Preparing teaching assistants to facilitate course-based undergraduate research experiences (CUREs) in the biological sciences: A call to action. CBE Life Sci. Educ. 22:es4.

Singh E., Vasishta P., Singla A. 2025. AI-enhanced education: exploring the impact of AI literacy on generation Z's academic performance in Northern India. Qual. Assur. Educ. 33:185–202.

Soto A.M., Weldon J.E., Hancock S.P. 2025. Faculty rewards from course-based undergraduate research experiences (CURE) in biochemistry. J. Microbiol. Biol. Educ. 26:e0016524.

Spell R.M., Guinan J.A., Miller K.R., Beck C.W. 2014. Redefining authentic research experiences in introductory biology laboratories and barriers to their implementation. CBE Life Sci. Educ. 13:102–110.

Suh W. 2025. Generative AI integration in higher education shifts students' attitudes from tool use to innovation. Discov. Educ. 4.

Tasdelen O., Bodemer D. 2025. Generative AI in the classroom: Effects of context-personalized learning material and tasks on motivation and performance. Int. J. Artif. Intell. Educ. 35:3049–3070.

Thorp H.H. 2026. Resisting AI slop. Science. 391:5.

Tian J., Zhang R. 2025. Learners' AI dependence and critical thinking: The psychological mechanism of fatigue and the social buffering role of AI literacy. Acta Psychol. (Amst.). 260:105725.

Townsend J.P., Hassler H.B., Dornburg A. 2023. Infection by SARS-CoV-2 with alternate

frequencies of mRNA vaccine boosting. J. Med. Virol. 95:e28461.

Townsend J.P., Hassler H.B., Dornburg A. 2025. Optimal annual COVID-19 vaccine boosting dates following previous booster vaccination or breakthrough infection. Clin. Infect. Dis. 80:316–322.

Townsend J.P., Hassler H.B., Sah P., Galvani A.P., Dornburg A. 2022. The durability of natural infection and vaccine-induced immunity against future infection by SARS-CoV-2. Proc. Natl. Acad. Sci. U. S. A. 119:e2204336119.

Townsend J.P., Hassler H.B., Wang Z., Miura S., Singh J., Kumar S., Ruddle N.H., Galvani A.P., Dornburg A. 2021. The durability of immunity against reinfection by SARS-CoV-2: a comparative evolutionary study. Lancet Microbe. 2:e666–e675.

Tran M., Balasooriya C., Semmler C., Rhee J. 2025. Generative artificial intelligence: the “more knowledgeable other” in a social constructivist framework of medical education. NPJ Digit. Med. 8:430.

Tzirides A.O. (olnancy), Zapata G., Kastania N.P., Saini A.K., Castro V., Ismael S.A., You Y.-L., Santos T.A. dos, Searsmith D., O’Brien C., Cope B., Kalantzis M. 2024. Combining human and artificial intelligence for enhanced AI literacy in higher education. Comput. Educ. Open. 6:100184.

Vygotsky L.S. 1978. Mind in society: the Development of Higher Psychological Processes. London, England: Harvard University Press.

Wang H., Dang A., Wu Z., Mac S. 2024. Generative AI in higher education: Seeing ChatGPT through universities’ policies, resources, and guidelines. Computers and Education: Artificial Intelligence. 7:100326.

Wang J., Davin K.J., Kissau S., Dornburg A. 2026a. From crisis to catalyst: How U.S. world language teachers are leveraging GenAI to navigate systemic challenges. Foreign Lang. Ann.

Wang Y., Chen G., Ma L., Ni J., Chen Q. 2026b. How does the mentor guidance function affect graduate students’ scientific research and innovation ability? Mediating role of scientific research self-efficacy. BMC Psychol. 14.

Ward J.R., Clarke H.D., Horton J.L. 2014. Effects of a research-infused botanical curriculum on undergraduates’ content knowledge, STEM competencies, and attitudes toward plant sciences. CBE Life Sci. Educ. 13:387–396.

Webb S., Massey D., Goggans M., Flajole K. 2019. Thirty‐five years of the gradual release of responsibility: Scaffolding toward complex and responsive teaching. Read. Teach. 73:75–83.

Wei Y., Perkins M. 2026. Generative AI and student collaboration: a scoping review of group work processes, outcomes, and risks. Int. J. Educ. Integr. 22.

Werth A., Oliver K., West C.G., Lewandowski H.J. 2022. Assessing student engagement with teamwork in an online, large-enrollment course-based undergraduate research experience in physics. Phys. Rev. Phys. Educ. Res. 18.

Weston M., Long T. 2026. How do biology undergraduates use AI-enabled feedback to revise scientific writing? Research Square.

Xiao J., Alibakhshi G., Zamanpour A., Zarei M.A., Sherafat S., Behzadpoor S.-F. 2024. How AI literacy affects students' educational attainment in online learning: Testing a structural equation model in higher education context. Int. Rev. Res. Open Distrib. Learn. 25:179–198.

Yan W., Nakajima T., Sawada R. 2025. Beyond tool use: Tracking the evolution of generative AI literacy among university students through a process-oriented investigation. Computers and Education: Artificial Intelligence. 9:100465.

Yilmaz R., Karaoglan Yilmaz F.G. 2023. The effect of generative artificial intelligence (AI)-based tool use on students' computational thinking skills, programming self-efficacy and motivation. Computers and Education: Artificial Intelligence. 4:100147.

Yoder E.B., Parker E., Frédérich B., Tew A., Jones C.D., Dornburg A. 2025. Multiple pathways of visual adaptations for water column usage in an Antarctic adaptive radiation. Ecol. Evol. 15:e70867.

Zhai C., Wibowo S., Li L.D. 2024. The effects of over-reliance on AI dialogue systems on students' cognitive abilities: a systematic review. Smart Learn. Environ. 11.

Zhu X., Wholey D.R. 2018. Expertise redundancy, transactive memory, and team performance in interdisciplinary care teams. Health Serv. Res. 53:4921–4942.